# Efficient fine-grained road segmentation using superpixel-based CNN and CRF models

1st Farnoush Zohourian
dept. Computer Science and Applied Cognitive Science
University of Duisburg-Essen
Duisburg, Germany
farnoush.zohourian@stud.uni-due.de

2nd Jan Siegemund
Delphi Electronics & Saftey
Delphi
Wuppertal, Germany
jan.siegemund@delphi.com

3rd Mirko Meuter
Delphi Electronics & Saftey
Delphi
Wuppertal, Germany
Mirko.Meuter@delphi.com

4th Josef Pauli
dept. Computer Science and Applied Cognitive Science
University of Duisburg-Essen
Duisburg, Germany
josef.pauli@uni-due.de

*Abstract*—Towards a safe and comfortable driving, road scene segmentation is a rudimentary problem in camera-based advanced driver assistance systems (ADAS). Despite of the great achievement of Convolutional Neural Networks (CNN) for semantic segmentation task, the high computational efforts of CNN-based methods is still a challenging area. In recent work, we proposed a novel approach to utilize the advantages of CNN's for the task of road segmentation at reasonable computational effort. The runtime benefits from using irregular superpixels as basis for the input for the CNN rather than the image grid, which tremendously reduces the input size. Although, this method achieved remarkable low computational time in both training and testing phases, the lower resolution of the superpixel domain yields naturally lower accuracy compared to high cost state of the art methods. In this work, we focus on an refinement of the road segmentation utilizing a Conditional Random Field (CRF). The refinement procedure is limited to the superpixels touching the predicted road boundary to keep the additional computational effort low. Reducing the input to the super-pixel domain allows the CNN's structure to stay small and efficient to compute while keeping the advantage of convolutional layers and makes them eligible for ADAS. Applying CRF compensates the trade-off between accuracy and computational efficiency. The proposed system obtains comparable performance among the top-performing algorithms on the KITTI road benchmark and it's fast inference makes it particularly suitable for real-time applications.

## I. INTRODUCTION

Development of intelligent vehicle and advanced driver assistance systems (ADAS) is one of the most active research areas and has attracted much attention recently. Camera-based perception of the drivable road environment is an important problem in this context. The variations in the illumination and appearance, apart from occlusions are challenging issues that make accurate road segmentation a difficult task. One solution to this is semantic segmentation, that assigns each pixel a category label. The breakthrough techniques such as deep convolutional neural network (CNN) models [Krizhevsky et al., 2012], [Simonyan and Zisserman, 2014] significantly improved pixel-wise semantic segmentation tasks by extracting rich hierarchical features [Long et al., 2015], [Lin et al., 2016]. However, fast and accurate estimation of the pixel labels in a way compatible for embedding into real-time application is not an easy and straightforward task. While accuracy of recent approaches is increased by creating deeper networks with as many layers as possible [Simonyan and Zisserman, 2014], [He et al., 2016], in practice, most of them are fairly limited in computational power and memory.

In our previous work [Zohourian et al., ], we proposed a novel approach to utilize the advantages of CNN's for the task of road segmentation at suitable computational effort. This method mainly differs from usual semantic segmentation methods in two aspects: first the input data model provided for the CNN network and second the simple CNN- network layering. The state of the art convolutional networks for image segmentation are based on two different input data model: They are based on either *patch-wise* or taking *full image resolution*. Most recent improvement in CNN are accomplished by using above input data mode and increasing the network size which require powerful GPUs. As deeper networks provoke large computational costs they are generally not suitable for embedded devices in self-driving cars and ADAS. In our work, the runtime benefits from using few layers and irregular superpixels as basis for the input for the CNN rather than regular *Patch* or *full image*, which tremendously reduces the input size. This strategy disassembles the pixel grid into superpixels forming the basic units for a pre-classification via a CNN. Reducing the input to the super-pixel domain allows the CNN's structure to stay small and efficient to compute while keeping the advantage of convolutional layers. Although, this method achieved remarkable low computational time in both training and testing phases, the lower resolution of the superpixel domain yields naturally lower accuracy compared to high cost state of the art methods.

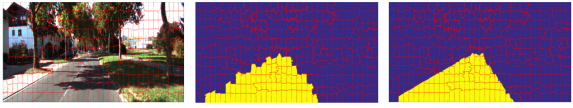

Fig. 1. a) Original image overlay with super pixel segmentation, b) CNN segmentation result based on supper pixel, c) CRF segmentation result.

The current work proposes a model strategy to refine the classification result from superpixel grid to pixel grid using Conditional Random Fields (CRFs) [Lafferty et al., 2001]. CRFs can model global properties like object connectivity, geometric properties and spatial relationship between objects (See Fig. 1). This work comprises two aspects for coupling local and global evidences. We combine the local image classification information extracted from CNN part with global information of neighboring pixel relations to decide for an accurate pixel label. The key idea of CRF inference for semantic labeling is to formulate the label assignment problem as a probabilistic inference problem that incorporates assumptions such as the label agreement between similar pixels or image regions. This idea follows largely previous work by applying CRF technique on CNN as a Post-processing step [Chen et al., 2016]. Our segmentation algorithm follows a number of steps which we briefly summarize here:

(a) segmenting the image into super-pixels, wherein the super-pixels are coherent image regions comprising a plurality of pixels having similar image features.
(b) determining image descriptors for the super-pixels, wherein each image descriptor comprises a plurality of image features.
(c) The super-pixels are assigned to corresponding positions of a regular grid structure extending across the image in order to create neighborhood relations for convoloutional purpose.
(d) This lattice together with the image descriptors are fed to the convolutional network based on the assignment to classify the super-pixels of the image according to semantic categories.
(e) An optimization strategy based on CRF refines the superpixel-based results to a pixel-based result to increase the precision of road segmentation.

Steps $a$ to $d$ are done in our previous work and the current work discusses the step $e$ (See Fig. 2). To keep the advantage in computation time, we limit the refinement scope to the superpixels bordered to street boundary estimated in the first step, i.e. neighboring superpixels assigned to a different class by the CNN result. Using CRF compensate the trade-off between accuracy and computational cost. The proposed system obtained comparable performance among the top-performing algorithms on the KITTI [Fritsch et al., 2013] road benchmark and its fast inference makes it particularly suitable for deployment in ADAS.

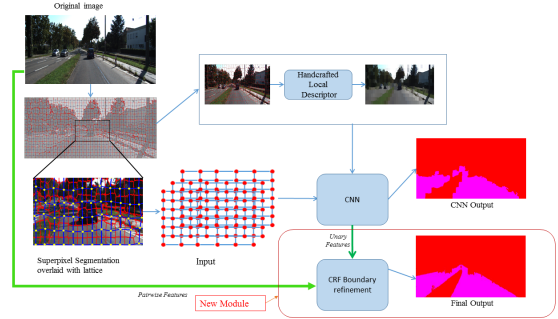

Fig. 2. The Proposed refinement SP-CNN architecture for road segmentation: Super pixels are extracted from the input image. Then a high dimensional local descriptor is computed for each superpixel. Finally, the superpixel descriptors assigned to the respective node in the lattice projection are fed into a CNN to classify the superpixels according to the semantic categories. A CRF technique is applied on CNN output to refine the segmentation result with respect to the native image grid.

## II. METHODOLOGY

In this paper, we combine both CRF and CNN into a framework to get a fine-grained road segmentation from urban scene images. Figure 2 displays the architecture of our method. First, we briefly explain our previously proposed method that provided a rational feature model fed into designed CNN network to segment road regions. The input data model is a combination of a higher dimensional feature space with irregular superpixel segmentations projected on a regular lattice structure, for convolutional purpose [Zohourian et al., ]. Afterwards, we explain how we improve the segmentation results by applying a CRF technique. The refinement procedure is limited to the superpixels touching the predicted road boundary. Restricting to this area helps to enhance the segmentation accuracy while keeping the additional computational effort low. To obtain the highest performance, we evaluated three different CRF techniques. The individual steps are presented in the next discussions.

### A. Superpixel-based Convolutional Neural Network

Superpixel segmentation is local grouping of pixels on the pixel grid, based on features like color, brightness, texture, etc [Ren and Malik, 2003]. Compared to pixel units, superpixel units store more compact information that can greatly reduce the model complexity and computation cost especially for real time systems. Well-segmented superpixels preserve the object structures and adjust well to the object contours, that causes the accuracy improvement of subsequent tasks like semantic segmentation.

To convolve the input data with kernels in convolutional layers of CNN, we need a regular structure (grid format). Irregular superpixels with different sizes or disordered shaped boundaries are not directly convolable, due to arbitrary neighborhood relations. On the other hand, enforcing a distinct topological structure of superpixel segmentation mostly prevent the maximum pattern homogeneity inside each superpixel.

In [Zohourian et al., ], we proposed a novel approach to have both maximum homogeneity and convolutional ability. First, we segment the image into coherent superpixels, comprising a plurality of pixels having similar image features. Then, for each irregular superpixel, an image descriptor is defined, which comprises a plurality of image features. The necessity of having a regular topology to be able to convolve the input data with kernels, motivated us to propose a superpixel lattice projection. The superpixels are projected into corresponding regular grid structure extending across the image. Finally, this lattice together with the image descriptors are fed to a convolutional neural network for pixel-wise classification purpose.

*1) Input Data Model:*
For superpixel segmentation, we used a modified version of SLIC algorithm [Achanta et al., 2010]. SLIC is adopted k-means clustering for grouping of pixels in the 5-D space defined by 3-D spectral space and 2-D spatial space. To enforce all superpixels to be connected and to avoid single isolated superpixels, SLIC applies an "Enforce-Connectivity" procedure which leads to non constant numbers of created superpixels in each iteration of adopted K-mean clustering making them unsuitable as direct CNN input model. To prevent this problem, in our modified version, we do not remove any small region with a single connected component. First, all label-connected components and adjacent superpixels in 2-D for each superpixel should be computed. Then for each superpixel with more than one segment and same label, we keep the larger one and merge the rest into the nearest superpixel. The nearest neighborhood is selected based on the euclidean distance between the center of sub-segment to the center of each adjacent segment. Now, we define our lattice centered in the rectangular structure extracted from the first iteration of SLIC method projected to the corresponding irregular superpixel from final step.

For having model relevant object characteristics in the image that are non-redundant and informative enough and facilitate the subsequent learning and generalization steps, we define for each superpixel a high dimensional feature descriptor. Each of the image descriptors comprises 69 image features consisting of 9 different color channels, 1 position and 59 Local Binary patterns(LBP). This provides high accuracy and reliability. The provided data model is fed to a simple convolutional network presented in the following.

*2) CNN Network Architecture:*
Due to our input data model with informative structure coming from superpixels and feature descriptor, our proposed method does not require a complex network architecture to handle large image context which leads to a considerable reduction of computational time. The network has two convolutional layers, two fully connected layers and one drop-out layer with non-linear activation function after each convolutional and fully connected layer. The input of our method is defined by the superpixel lattice on each image with size of $H/S$ and $W/S$, where S is the initial superpixel size and $W, H$ are image width and hight. The output is a set of three numbers to indicate which of the three classes of the *road*, *non-road* or *un-labeled* they belong to [Zohourian et al., ].

*B. Segmentation Refinement with CRF*

Our superpixel-based convolutional network already has the capability of modeling global relationships within a scene and to adaptively coarse segmentation by capturing the local properties such as object shape and contextual information. However, CNNs have their shortcomings to model the interactions and correlation between the output variables directly, which is important for a smooth semantic segmentation. To this end, graphical models like Conditional Random Fields (CRFs) have been used to impose consistency and coherency between labels. CRFs can model global properties like object connectivity, geometric properties and spatial relationship between objects. By combining CNN and CRF models, we are able to fine-tune the CNN segmentation results specially along the road border. CRFs are formulated in a probabilistic framework, modeling the joint probability of the image and its corresponding labels. We represented CRF in terms of energy minimization, where the energy function has two terms: one term penalizes solutions that are inconsistent with the observed data, called Unary term ($\varphi$), while the other term enforces some kind of spatial coherence, called pairwise term ($\psi$) (See Eq.1).

$$E(l,I) = \alpha \sum_i \varphi(l_i|x_i) + \beta \sum_{ij} \psi(l_i,l_j|x_i,x_j) \quad (1)$$

where $l_i$ is corresponding class label, $I$ is $\{x_1,...,x_N\}$, and $x_i$ is the pixel intensity. The unary potential is the inverse likelihood of each pixel getting a particular label. The pairwise part encodes the neighboring information and computes the difference between a pixel label from its neighbor.

In this work, we evaluated three inference methods for pixel-wise image segmentation, considering the labels of the image as hidden states and solving the label prediction as a solution of the maximum posterior probability (MAP) [Nowozin et al., 2011]. We investigate quality of all three methods for classification and study the influence of computational gain and overall accuracy, allowing for robust and accurate statistical analysis on road segmentation. Image data is used to build the CRF that potential functions which are applied directly on image pixels belong to the super-pixels touching the road boundary. In all three inference methods, the respective unaries are supplied by CNN. The three inference methods are:

*1) Iterated Conditional Modes (ICM):*
Iterated Conditional Modes (ICM) [Besag, 1986] is an iterative algorithm that employs a deterministic greedy strategy to find a local minimum by minimizing an energy function (See Algorithm (1)). We initiated ICM by projecting the CNN prediction from superpixel to pixel grid. We smooth out the initial segmentation by assigning the new label to each pixel. This process is repeated until convergence. To estimate appropriate values for the variances of the pixel intensity levels, forming the unary term, we used Gaussian Mixture Model [Williams and Rasmussen, 1996]. GMM has 3 components belonging to either road and non-road classes and is modeled as two separate Gaussian mixture models. Unary term in formula 1 is extended to $\alpha \sum_i \varphi(l_i, k_i, \theta|x_i)$ Where $k_i \in \{1,...,K\}$ are the Gaussian mixture components, $k$ is number of classes and $\theta$ is GMM Model parameters. GMM parameters are learned by K-mean method as an iterative Expectation-Maximization (EM) algorithm.

In binary term, the neighborhood is based on 4-neighbors in pixel grid. For each class, we compute the sum of the labels which are not equal to the current pixel label and penalize the total energy of each class with some threshold value.

---

**Algorithm 1:** Updated ICM

ICM $(I, L, T, MaxItr)$
**Input** : pixels contained in superpixels surrounding the road border $(I)$, corresponding labels from CNN $(L)$, Potential threshold $(T)$, maximum number of iterations $(MaxItr)$
**Output:** smooth road segmentation
Initialize Mixture Models with CNN outputs          // GMM
Assign GMM components
$$K_i = \arg\min_k \varphi(l_i, k_i, \theta|X_i)$$
Learn GMM parameters
$$\theta = \arg\min \sum_i (l_i, k_i, \theta|X_i)$$
Calculate unary energy          // Unary part
$$\varphi(l_i, k_i, \theta|X_i) = -\log \pi(l_i, k_i) + \frac{1}{2}\log|\sum k_i|$$
$$+ \frac{1}{2}(l_i - \mu(k_i))^T \sum k_i (l_i - \mu(k_i))$$
**while** *not MaxItr* **do**
  Calculate pairwise energy          // Pairwise part
  $$\psi(l_i, l_j|x_i, x_j) = \begin{cases} x_j & \text{if } l_i \neq l_j, \text{ j\&i are neighbors} \\ 0 & \text{otherwise} \end{cases}$$
  Predict labels
  $$l = \arg\min \mathbf{E}(l, k, \theta|I)$$
**end**

---

*2) Loopy Belief Propagation (LBP):*
For a more accurate message passing, we apply Loopy Belief Propagation (LBP) [Murphy et al., 1999] on the same energy term. We solved LBP by min-Sum algorithm [Kschischang, 1999]. As for ICM method, the input to our LBP method are all pixels that belong to the super pixels which touch the road border with corresponding label estimated from CNN output. The efficient approximate inference requires the Gaussian kernels computed over elements. The Gaussian Mixture Model with 3 components for each road and non-road classes is used. LBP is an iterative algorithm and will terminate if the changes in energy drops below a threshold or after a fixed number of iterations. At each iteration messages are passed around the MRF model. Our choice of the message passing scheme is right, left, up and down. Once LBP iteration completes, the label with highest energy is assigned to each pixel.

---

**Algorithm 2:** Modified LBP (Min-Sum)

LBP $(I, L, T, MaxItr)$;
**Input** : pixels surrounded the road border $(I)$, corresponding labels from CNN $(L)$, convergence tolerance $(T)$, maximum number of iterations $(MaxItr)$
**Output:** approximate MAP labeling
Initialize Mixture Models with CNN outputs;          // GMM
Assign GMM components (see Algorithm 1)
**while** *not MaxItr* **do**
  Update Message;          // Min-Sum method
  $$msg_{i \to j}(l) = \min_{l' \in \text{all labels}} [DataCost(x_i, l') +$$
  $$smoothnessCost(l, l') +$$
  $$\sum_{k=\text{neighbors of i expect j}} msg_{k \to i}(l')]$$
  where $msg_{i \to j}(l)$ is the message from node $i$ to node $j$ for label $l$. Calculate Belief
  $$Belief(z_i = l) = DataCost(x_i, l)) +$$
  $$\sum_{k=\text{neighbors of i}} msg_{k \to i}(l)$$
**end** MAP assignment and calculating the energy;

---

*3) Dense CRF with Gaussian edge potentials:*
Contrary to the previous inference methods, we used different energy term here. The unary potential is obtained by the class conditional probability map obtained from soft-max function of CNN network for those superpixels that touched the road border. This is given to the fully-connected CRF proposed by [Krähenbühl and Koltun, 2011] for the pixel-wise labeling. The pairwise potentials are usually modeled based on the relationship among neighboring pixels and weighted by color similarity, whereas dense CRF model considers long range interactions among pixels instead of just neighboring information by using a fully-connected graph, where all pairs of image pixels, $i, j$ are connected together.

The pairwise potential is defined based on formula 2 where the first term depends on both pixel positions ($P_i$ and $P_j$) and pixel color intensities ($I_i$ and $I_j$) and the second term only depends on pixel positions. $\omega_m$ ($m = \{1, 2\}$) are linear combination weights. The terms are defined as Gaussian kernels whereby, first represents the color-similarity between nearby pixels and the latter removes all small isolated regions. The degrees of nearness and similarity are controlled by parameters $\sigma_\alpha$ and $\sigma_\beta$ and the size of small areas is thresholded by $\sigma_\gamma$. The algorithm uses the mean-field approximation and a message passing scheme in a fully-connected graph to efficiently infer the latent variables that approximately minimize the Gibbs energy of a labeling [Krähenbühl and Koltun, 2011].

$$\psi_{i,j} = \omega_1 \exp(-\frac{\|P_i - P_j\|^2}{2\sigma_\alpha^2}) - \frac{\|I_i - I_j\|^2}{2\sigma_\beta^2}) + \omega_2 \exp(-\frac{\|P_i - P_j\|^2}{2\sigma_\gamma^2}) \quad (2)$$

## III. EXPERIMENTS AND RESULTS

We evaluate our method on public KITTI [Fritsch et al., 2013] dataset comprising urban scenarios. KITTI comprises 502 8-bits RGB images splits in train, validation and test sets with ground truth label for three semantic classes. The training set has 289 images (95 images with urban markings (UM), 96 images with multiple urban markings (UMM) and 98 images where the street has no urban markings (UU). The test set has 290 images including (96 UM, 94 UMM and 100 UU) images. The image dimensions are different in the width lying in [1226, 1238, 1241, 1242] and the height in [370, 374, 375, 376]. We selected 20% of training set images from 3 different categories UM,UMM,UU for the validation set. These images completely originate from different video sequences which are not part of the training set.

For having a fair comparison between our current approach and our previous work [Zohourian et al., ], we use the same conditions and parameter values. We used the hardware specifications, CPU: Intel(R) Core(TM) i7-4790K @4GHz for both training and testing. In [Zohourian et al., ] we used SLIC parameters $K = 400, m = 35$ resulting in 396 super pixels in each image projected to a $11 \times 36$ lattice for CNN input. Our original task was to segment road from non-road (background). We evaluated the accuracy once on pixel and once on superpixel level. For evaluation in the superpixel domain, the ground truth for each Super-pixel is defined based on the majority pixel-labels inside the superpixel.

In current work, We evaluate the performance of the proposed approach implemented in three different CRF techniques with the accuracy of the pixel grid obtained from superpixel-based convolutional network [Zohourian et al., ] based on the image perspective and a birds eye projection provided by KITTI dataset.

| Method | ACC | F1 | PRE | REC | FPR | FNR |
|---|---|---|---|---|---|---|
| CNN classifier | 94.41% | 95.18% | 91.41% | 97.34% | 8.59% | 2.60% |
| CRF_LBP | 94.31 % | 87.58 % | 90.49 % | 86.74 % | 9.51 % | 4.92% |
| CRF_ICM | 96.90 % | 87.94 % | 83.05 % | 93.62 % | 16.95 % | 0.89 % |
| CRF_Meanfeild | **96.85** % | **90.94** % | 92.30 % | 90.65 % | 7.70 % | 2.10% |

TABLE I
EVALUATION RESULTS ON KITTI VALIDATION SET BEFORE AND AFTER APPLYING CRF TECHNIQUES

| Benchmark | MaxF | AP | PRE | REC | FPR | FNR |
|---|---|---|---|---|---|---|
| UM_ROAD | 83.22 % | 72.94 % | 77.11 % | 90.39 % | 12.23 % | 9.61 % |
| UMM_ROAD | 90.96 % | 84.63 % | 87.86 % | 94.29 % | 14.32 % | 5.71 % |
| UU_ROAD | 80.02 % | 67.93 % | 77.56 % | 82.64 % | 7.79 % | 17.36 % |
| URBAN_ROAD | 85.97 % | 77.81 % | 82.04 % | 90.31 % | 10.89 % | 9.69 % |

TABLE II
EVALUATION RESULTS ON KITTI TEST SET BASED ON MEAN-FIELD CRF

| Method | Processor | MaxF | AP | Runtime(s) |
|---|---|---|---|---|
| LODNN | NVIDIA GTX980Ti GPU, 6GB memory | 96.05 % | 95.03 % | 0.018 |
| UP_CONV_POLY | NVIDIA Titan X GPU. | 95.52 % | 92.86 | 0.083 |
| DDN | NVIDIA GTX980Ti GPU, 6GB memory | 94.17 % | 92.70 % | 2 |
| **Ours** (without CRF) | Intel(R) Core(TM) i7-4790K CPU @4GHz | 85.07 % | 79.86 % | **0.019** |
| **Ours** (With CRF) | Intel(R) Core(TM) i7-3770 CPU @ 3.40GHz | 90.96 % | 84.63 % | 0.21 |

TABLE III
KITTI ROAD SEGMENTATION PERFORMANCE (IN %) ON URBAN MULTIPLE MARKED (UMM) CATEGORY. ONLY RESULTS OF PUBLISHED METHODS ARE REPORTED. LODNN: [CALTAGIRONE ET AL., 2017], UP_CONV_POLY [OLIVEIRA ET AL., 2016], DDN [MOHAN, 2014]

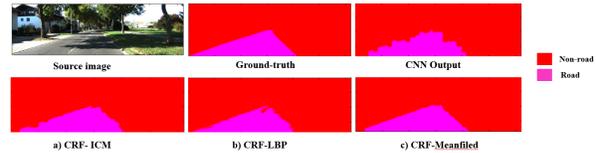

Fig. 3. Road segmentation based on a) Iterated conditional Mode,b) Loopy Belief Propagation, c)Fully connected CRf with mean-field approximation method

### A. Evaluation on Image Perspective

The ground truth for the test data of the KITTI data set is not publicly available (See [Fritsch et al., 2013]), hence we used the validation set to be able to evaluate our approach on the image perspective. First, we evaluated our result obtained from all three proposed CRF techniques on validation set. We followed the evaluation scheme presented in the KITTI dataset (See [Fritsch et al., 2013]). Table I summarizes the average evaluation results on the validation set for all urban categories. The accuracy obtained from CNN part was 94.41%. In the current approach by applying CRF technique we have around 2% improvement in the accuracy and we could reach about 96.85%. Fully connected CRF based on mean-filed approximation has the best performance. We fix the number of mean field iterations to 20. The appearance kernel weight is $\omega_1 = 0.1$ and the kernel widths are $\sigma_\alpha = 60, \sigma_\beta = 10$. The smoothness kernel parameters are assigned $\omega_2 = 3, \sigma_\gamma = 1$. Iterated conditional Mode (ICM) with slightly lower accuracy is ranked second. The empirical evaluation yields the most suitable potential weight $\alpha = 1$ and $\beta = 20$ (See Equ.1). Loopy belief propagation acquired worst results. We obtained the represented results with the potential weight $\alpha = \beta = 1$. For all three methods, we fix the number of iteration to 20. Figure 3 shows one representative result based on all three CRF techniques.

### B. Evaluation on Birds Eye Perspective

For evaluation in birds-eye perspective in the KITTI benchmark the images are projected on the ground plane via the known camera geometry. We had an improvement of approximately 5% on official KITTI Test set compared to our CNN classifier without CRF refinement (See Table.III). We could fix the weaker accuracy in the birds eye view projection compared to the image perspective evaluation, which appeared due to spreads of the error induced by inaccurate superpixels on the road border over a much larger region [Zohourian et al., ]. Table II shows the results which are split into the different road types (UM, UMM, UU, URBAN). Two samples in BEV are shown in Figure 4. Whilst the street is nicely segmented, there are a few false detections that mostly happened when the segmented area was fooled by a shadow covering the street. Moreover, compared to the results obtained from validation set we still have lower average accuracy on KITTI test set, which contains more challenging and diverse scenarios.

5
516Copyright © 2018 by CENPARMI, Concordia University

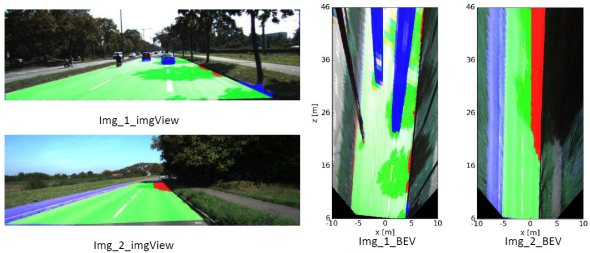

Fig. 4. Road segmentation result from official Kitti test set in baseline and bird eye view perspectives. Here, red denotes false negatives, blue is false positives and green represents true positives.

*C. Run-time Analysis*

Most of the state of the art methods in semantic segmentation are based on GPU and powerful hardware facilities which limit their application to CPU based embedded systems. Using superpixels and simple CNN network combined with optimized pixel-wise CRF technique which is applied only on the small portion of pixels surrounding the road contour distinctly reduces the computational complexity. In Table III we compared our approach with some of the state of the art methods in semantic segmentation in both accuracy and computational time. All results are provided in KITTI *URBAHN Multi-line ROAD* test set. The state-of-the-art method [Caltagirone et al., 2017] has a run-time $0.018s$, but their method is implemented in torch and uses NVIDIA GTX980Ti GPU with 6GB memory, whereas our proposed method uses 4core CPU. Experiments conducted on KITTI ROAD dataset required only $4.3s$ training time per epoch for one image ( [Zohourian et al., ]). For testing the total runtime of our approach amounts to $0.019s$ per image including SP segmentation, Feature extraction and CNN based on CPU specification. Our unoptimized Matlab implementation of mean-filed CRF requires almost $0.2s$.

To sum up, we can emphasize that our approach is compatible for real-time systems with a reasonable trade-off between accuracy and timing cost based on very cheap hardware facilities that make it very fast in both training and testing parts.

## IV. CONCLUSION AND FUTURE IMPROVEMENTS

We proposed an approach to refine the segmentation maps obtained from a super-pixel based convoluotional neural network applied for semantic road segmentation. We focused to improve the segmentation accuracy, especially at the road boundary. We formulate the problem using CRF technique which embeds the potential from a global CNN and a local regional CRF. Our main goal focuses on a strategy to reduce the runtime for a pixel-wise classification while still achieving a high level of accuracy to make our approach suitable for deployment on embedded devices for ADAS applications. Our algorithm achieves promising results in semantic road segmentation on KITTI dataset.

Future work will initially focus to improve the discrimination of the road pattern in challenging conditions; such as shadow on road surface, illumination changes or similarity with neighboring patterns like sidewalks. It is also promising to evaluate this approach for more than 2 classes and extend the pixel-wise classification to different objects such as sidewalks,lane, traffic sign, vehicles, buildings,sky, etc.